\documentclass[10pt,twocolumn,letterpaper]{article}

\usepackage{cvpr}
\usepackage{times}
\usepackage{epsfig}
\usepackage{graphicx}
\usepackage{amsmath}
\usepackage{amssymb}
\usepackage{color}
\usepackage{algorithm}
\usepackage{algpseudocode}
\usepackage{epstopdf}
\usepackage{subcaption}

\usepackage[pagebackref=true,breaklinks=true,letterpaper=true,colorlinks,bookmarks=false]{hyperref}

 \cvprfinalcopy 

\def\cvprPaperID{1166} 

\ifcvprfinal\fi
\begin{document}

\title{Collaborative Feature Learning from Social Media}



\author{
Chen Fang\textsuperscript{1}\hspace*{12mm} Hailin Jin\textsuperscript{2}\hspace*{12mm} Jianchao Yang\textsuperscript{3}\hspace*{12mm} Zhe Lin\textsuperscript{2}\\
\textsuperscript{1}Dartmouth College\hspace*{8mm} \textsuperscript{2}Adobe Research\hspace*{8mm} \textsuperscript{3}Snapchat\\
{\tt\small chenfang@cs.dartmouth.edu}\hspace*{1mm} {\tt\small \{hljin, zlin\}@adobe.com }\hspace*{1mm} {\tt\small jianchao.yang@snapchat.com}
}

\maketitle

\begin{abstract}
Image feature representation plays an essential role in image recognition and related tasks. The current state-of-the-art feature learning paradigm is supervised learning from labeled data. However, this paradigm requires large-scale category labels, which limits its applicability to domains where labels are hard to obtain. In this paper, we propose a new data-driven feature learning paradigm which does not rely on category labels. Instead, we learn from user behavior data collected on social media. Concretely, we use the image relationship discovered in the latent space from the user behavior data to guide the image feature learning. We collect a large-scale image and user behavior dataset from Behance.net. The dataset consists of 1.9 million images and over 300 million view records from 1.9 million users. We validate our feature learning paradigm on this dataset and find that the learned feature significantly outperforms the state-of-the-art image features in learning better image similarities. We also show that the learned feature performs competitively on various recognition benchmarks.
\end{abstract}

\section{Introduction}
\label{sec-intro}
Image recognition is a central problem in Computer Vision which enjoys great progresses in the last decade.
Feature learning plays an essential role in image recognition. 
Traditional recognition methods, such as \cite{1641019,lowe1999object,5206772,BergamoTorresani:CVPR2012,torresaniSF10,DBLP:journals/ijcv/LiSLL14}, are based on hand-crafted image features. 
These feature representations require a significant amount of domain
knowledge and do not generalize well to new domains. 
The current state-of-the-art feature learning paradigm is supervised learning from data \cite{DBLP:conf/nips/KrizhevskySH12}.
This data-driven supervised feature learning paradigm does not require domain knowledge but requires
large datasets with category labels to train properly.
However, collecting large labeled datasets is not an easy task even with the help of crowdsourcing.
For instance, we often need experts to label images in a special domain which may be hard to find via crowdsourcing.
Lack of large labeled datasets limits the applicability of the supervised feature learning paradigm in new problem domains.

\begin{figure}[t!]
\begin{center}
\includegraphics[width=\columnwidth]{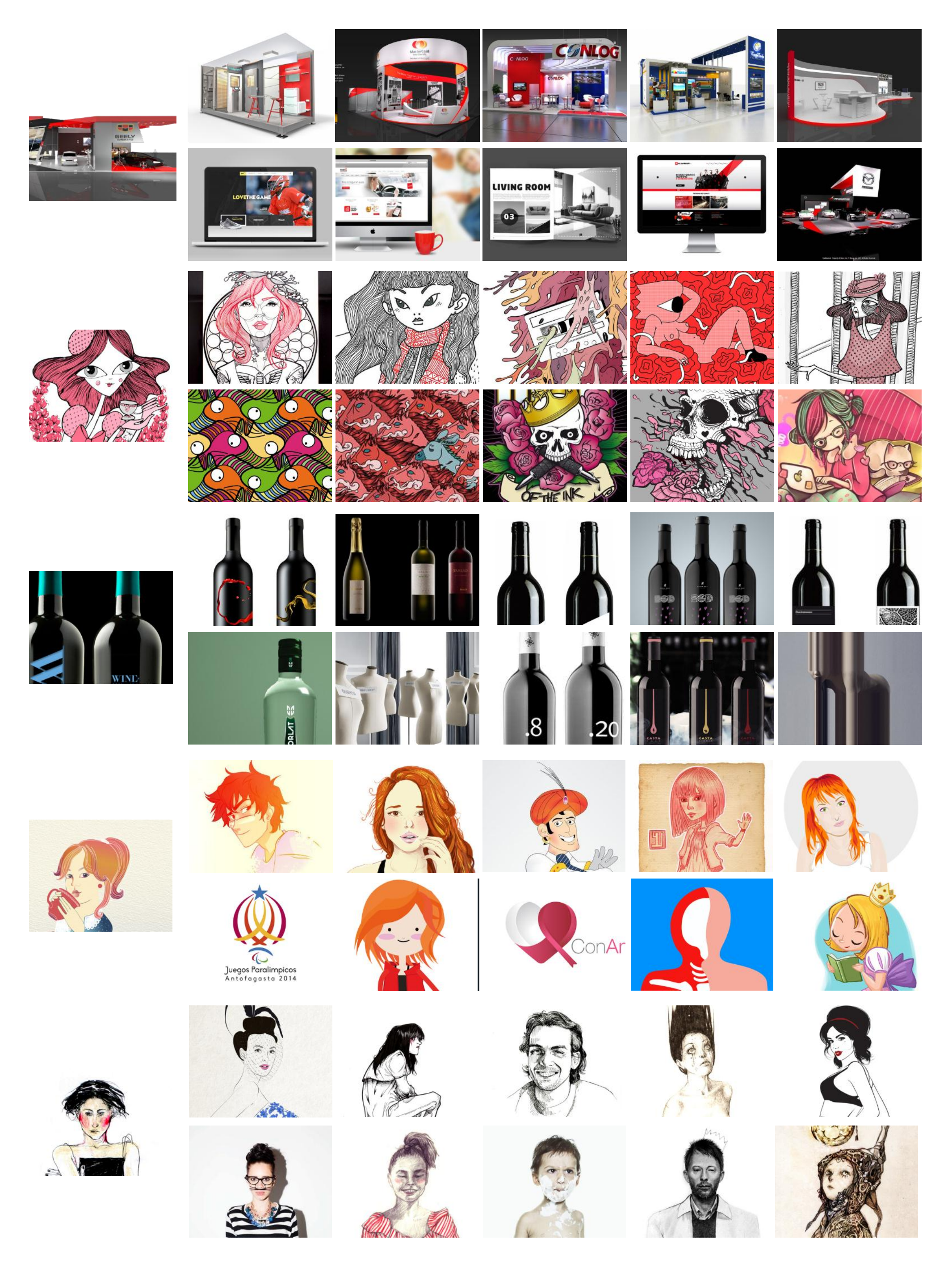}
\end{center}
\caption{Comparisons of our feature and the ImageNet feature
  \cite{DBLP:conf/nips/KrizhevskySH12} for image similarity on the Behance
  data. The leftmost images are the queries and the rest are the
  top-5 nearest neighbor images based on the cosine similarity of the
  features (no re-ranking). The top rows are from our feature and the
  bottom rows are from the ImageNet feature. One can observe that our
  feature yields more similar images in terms of both content and
  style.}
\label{fig-feature-compare}
\end{figure}

Since there are plenty of unlabeled image data, 
the general directions in the literature to overcome the dataset issue
include unsupervised feature learning \cite{daiV13,
  leeGRN09,DBLP:conf/cvpr/RanzatoHBL07,Raina07self-taughtlearning,DBLP:journals/corr/abs-1205-3137}
and transfer learning (using labeled data from different domains)
\cite{quattoniCD08}.  These methods hold great promise, but regarding
how successful they are, there remains an interesting research question: {\em
  Are category-level labels the only way for data driven feature
  learning?}

There is a surge of social media websites in the last ten years. 
Most social media websites such as Pinterest have been collecting
content data that the users share as well as behavior data of the users.
User behavior data are the activities of individual users, such as likes, comments, or view histories 
and they carry rich information about corresponding content data. 
For instance, two photos of a similar style on Pinterest tend to be
pinned by the same user. If we aggregate the user behavior data across
many users, we may recover interesting properties of the content. For
instance, the photos liked by a group of users of similar interests tend
to have very similar styles.

In this paper, we propose a new paradigm for data driven image feature learning which we
call {\em collaborative feature learning.}
The main idea in collaborative feature learning is to learn image
features from user behavior data on social media.
In particular, we use the user behavior data collected on social media 
to recover latent representations of individual images
and learn a feature transformation from the images to the recovered latent representations.
It is a major departure from the existing paradigms on feature learning
such as supervised learning in that we do not rely on category labels at all.
There are several challenges in this new feature learning paradigm. 
In particular, user behavior data can be very sparse and noisy. 
For instance, most users only see a very small portion of all the images
and some user behavior data are erroneous. 
Fortunately, there exist structures in the behavior data over all the users
and the structures can be exploited to deal with sparsity and noise in the data.

We acknowledge that our new data-driven feature learning paradigm only
applies where there are social media data available and the
effectiveness is determined by the quality of the data.  In fact, all
the data-driven methods share the same limitation.  To test our feature
learning paradigm, we collect a large-scale dataset from Behance.net
which is a popular social media that focuses on artists and
designers. We download about 1.9 million Behance artworks along with the
view history of about 1.9 million users. That results in more than 300
million user-artwork view records. In the experiments, we find that the
learned latent representations indeed reflect rich visual and semantic
information of the images. We further observe that the image features
learned from the latent representations not only perform well on
standard image recognition benchmarks but outperform the
state-of-the-art feature (the ImageNet feature
\cite{DBLP:conf/nips/KrizhevskySH12}) on tasks such as finding images of
similar styles (see Figure~\ref{fig-feature-compare}).

\subsection{Main contributions}
\label{sec-contrib}
We propose a new paradigm for feature learning from social media. We completely forgo the use of category labels in existing feature learning paradigms. Instead, we use user behavior data collected on social media. Our paradigm can take advantage of the massive data that are collected on social media which mitigate the dataset scalability issue in feature learning and image recognition in general. We further validate and test our paradigm on large-scale data collected from a real-world social media website, and show promising results. Finally, we want to remark that although the focus of this paper is image and visual data, our feature learning paradigm is by no means limited to learning visual features. For instance, it can be used to learn interesting audio features from social media websites such as Spotify~\cite{van2013deep}.


\subsection{Related work}
\label{sec-related}
Image features play an important role in various image recognition
problems. There is a rich body of literature in Computer Vision on image
features. It is beyond the scope of this paper to do a comprehensive
review. Early methods 
\cite{1641019,lowe1999object} use low-level
features which are more about
appearance and recent methods, such as
\cite{5206772,BergamoTorresani:CVPR2012,torresaniSF10,DBLP:journals/ijcv/LiSLL14}, focus on
high-level features which are more about semantics. Different from
hand-crafted features, features learned directly from data are the current
state-of-the-art \cite{DBLP:conf/nips/KrizhevskySH12}.  Data-driven features are
shown to be able to effectively encode both semantics and appearance and
outperform previous methods on many recognition benchmarks. But they
need a lot of labeled images (on the order of millions) to train
properly. Unsupervised feature learning methods, just to name a few
\cite{daiV13,
  leeGRN09,DBLP:conf/cvpr/RanzatoHBL07,Raina07self-taughtlearning,DBLP:journals/corr/abs-1205-3137},
hold significant promise in terms of overcoming the labeled dataset
limitation.

In terms of using social media data for learning, our work is related to
\cite{van2013deep} and \cite{6810890}.  \cite{van2013deep} addresses the
cold start problem in music recommendation.  It mines latent factors
from user music listening logs and uses convolutional neural networks as
a nonlinear regressor to predict the latent factors. Different
from~\cite{van2013deep}, this work focuses on how to learn image
features from user behavior data. This work is also different from
\cite{6810890} which uses deep belief nets to fuse and learn unified
features for multi-modal social media data. Instead of learning
features, \cite{6810890} starts from low-level features of multi-modal
social media data and focuses on learning a common feature map for various tasks.

Our method uses singular value decomposition based collaborative filtering
which is a well studied area in recommender systems
\cite{eigentaste}. In particular, we adopt the ideas in
\cite{DBLP:conf/icdm/HuKV08} to handle implicit feedback data and
combine them with the negative sampling strategy proposed in
\cite{DBLP:journals/jmlr/DrorKKW12}.

\section{Collaborative Feature Learning}


The proposed approach is a framework that unifies latent factor analysis and deep convolutional neural network for image feature learning from social media. Although rich social information can be harvested from social websites, such as content items, item tags, user social friendships, user views and comments, we focus on the simple form of user-item view data in this work to keep our feature learning framework general. Given a set of content items $\mathcal{I}=\{I_{1},\ldots,I_{M}\}$ and a set of users $\mathcal{U}=\{U_{1},\ldots,U_{N}\}$, the corresponding user-item view data is in the format of a matrix between $\mathcal{I}$ and $\mathcal{U}$, which is denoted as $V \in \mathbb{R}^{M \times N}$. We set $V_{ij} = 1$ if $U_j$ viewed content item $I_{i}$ (regardless the number of views), and $V_{ij} = -1$ otherwise (missing entries). Note that we use $-1$ to denote the missing entries, which should not be confused with negative signals mentioned in following sections.  As we will show later, the user-item view matrix encodes a lot of information about the similarity between different content items, which we can use for supervised image feature learning for the social content. The content items could be any media format presumably, such as videos, images, or audios. In this work, we will focus on images and leave feature learning for other media formats as future work.

\begin{figure}
\centering
\hspace*{-3mm}\includegraphics[scale=0.26]{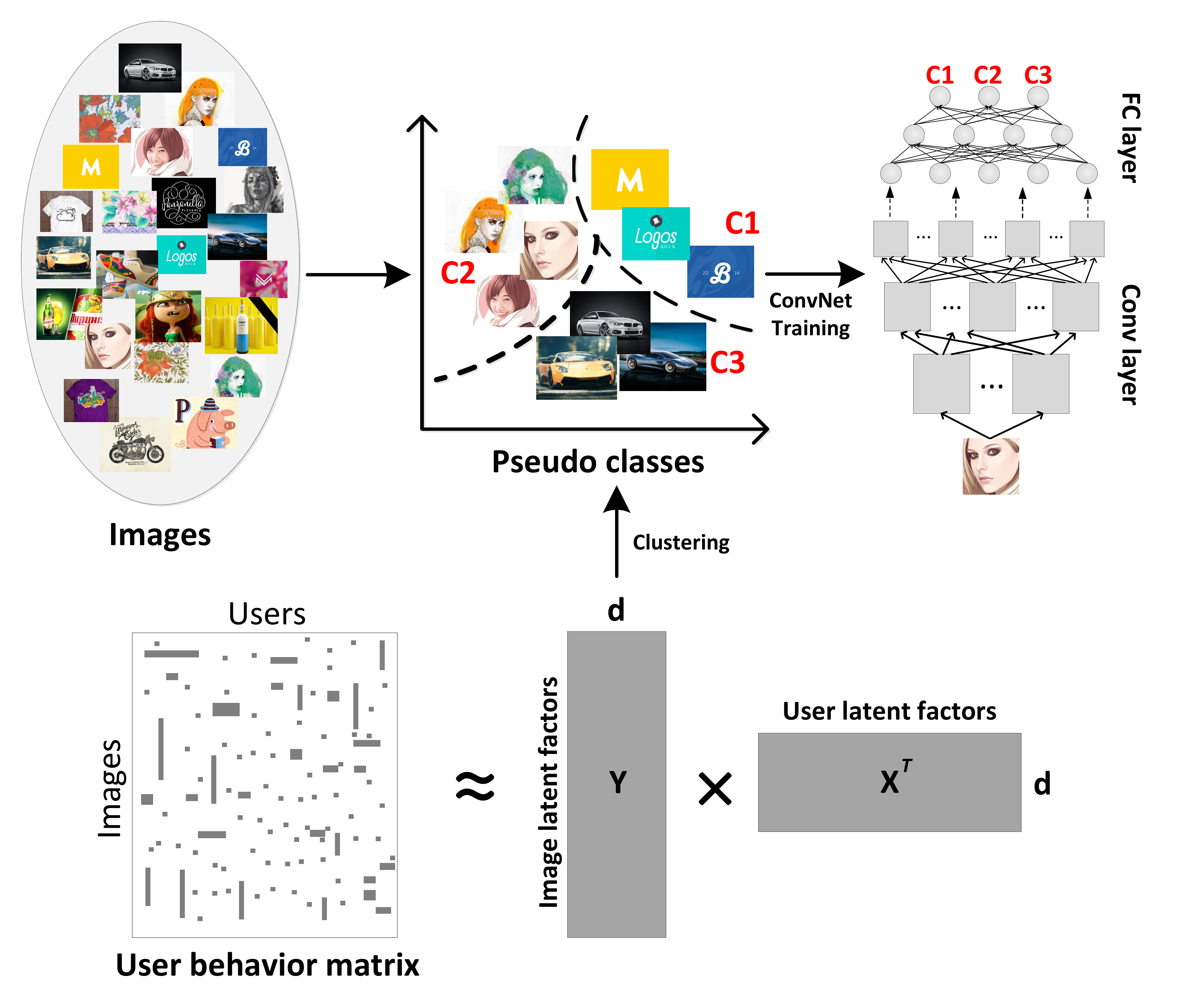}
\caption{Approach overview. }
\label{fig:framework}
\end{figure}

Figure~\ref{fig:framework} provides an overview of our approach. Based on the user-item view matrix, we use collaborative filtering to decompose it into the product between content item latent factors and user latent factors. As the latent factors of content items encode rich information about the similarity between the content items, we then generate pseudo classes for the content items by clustering their corresponding latent factors using K-means. Deep convolutional neural network (DCNN)~\cite{DBLP:conf/nips/KrizhevskySH12} is then trained based on these pseudo classes in the traditional supervised way. Finally, the trained DCNN can be used to extract content features for our social content domain. In the following sections, we will cover details of our approach.

\subsection{Collaborative filtering for latent factor analysis}

In this part, we explain how to extract content latent factors from the user-item view matrix.  
\paragraph{Implicit feedback data.} The view matrix we consider here only records the minimum information of whether a user viewed a particular content item or not. This is called \emph{implicit feedback} in the literature, which is an indirect reflection of a user's true opinion of the content item. Compared with explicit feedback data, such as user ratings on Netflix about movies and on Amazon about products, where explicit positive or negative feedbacks are given, implicit feedback data is more general and at a much larger scale. For example, Amazon can collect many more clicks and views than getting reviews or comments from the buyer in a much easier way. However, implicit feedback data is typically a lot noisier and weaker indication of the user's true opinion. Since it does not contain explicit negative signals, there is no easy way to identify negative signals from the missing data, because a missing entry could be interpreted as a sign of dislike or the user just has not discovered the content yet. Given the massive amount of contents in social media, the user-item view matrix is extremely sparse (e.g., well over 99\% entries are missing), and it is very likely that most of the missing entries are due to that the content has not been discovered by the user.  

To address this issue, inspired by the experience from an industrial competition on building music recommendation systems~\cite{DBLP:journals/jmlr/DrorKKW12}, we sample ``negatives'' from the large number of missing entries with a probabilistic trick. When drawing the negative samples, the sampling follows a probability distribution that is proportional to the popularity of the content. The rationale behind our sampling is that, a popular content has a higher chance of being discovered by the user, and therefore, a missing entry is more likely to suggest an negative altitude for the user. The popularity of a content item is a measure of how much exposure it receives from users, i.e., how many users have viewed it. Formally, given the view matrix $V$, we define the popularity $p_i$ of item $I_i$ as
\begin{equation}
p_{i} = \sum_{\{j: V_{ij} = 1\}} V_{ij}.
\end{equation}
Based on the content popularity, the sampling distribution for negative data is defined as:
\begin{eqnarray}
Pr_{ij} & \propto & 
                 \begin{cases}
                 {p_i},  & V_{ij} = -1\\
                 0, & V_{ij} = 1
                 \end{cases} 
\end{eqnarray}
Apparently, we avoid sampling entries that are positive. In practice, we
take the logarithm of $V_{ij}$, and normalize $Pr_{ij}$ with respect to
each user, so that for each user, the sampling probabilities sum up to
1. Algorithm~\ref{alg:sampling} provides a description of our sampling
process to get negative data, where $\mathcal{N}$ is the set of sampled
negative view entries. For every missing entry $V_{ij} \in \mathcal{N}$, we set $V_{ij} = 0$ for further analysis. 

\paragraph{Matrix factorization.} 
Among different methods in collaborative filtering for latent factor analysis, matrix-factorization-based models have gained a lot popularity 
thanks to their attractive accuracy and scalability. In this work, we will focus on using matrix factorization model on our user-item view matrix. 
The model associates each user $U_j$ with a user latent factor vector $x_j \in \mathbb{R}^d$ and each item $I_i$ with an item latent factor 
vector $y_i \in \mathbb{R}^d$, where $d \ll |\mathcal{I}| $ and $d \ll |\mathcal{U}|$ is the dimension of the latent space. And the prediction 
for an entry $V_{ij}$ is done by taking the inner product between the two latent factors, i.e., $\hat{V}_{ij} = y_i^T x_j$. With the positive 
entries and sampled negative entries from our user-item view matrix, we conduct matrix factorization with regularizations as
\begin{equation}
\label{eq:SVD}
\min_{x_*,y_*} \sum_{V_{ij} \neq -1}
(V_{ij}-y_i^Tx_j)^2+\lambda(||x_j||^2+||y_i||^2)
\end{equation}
Here $\lambda$ is the weight placed on the regularization term. Note that the summation is over ``non-missing'' entries only in $V$, including both positives and the sampled negatives. Due to the large size of our problem (on the magnitude of millions), we adopted stochastic gradient descent (SGD) to solve Equation (\ref{eq:SVD}). 
At each iteration of SGD, a single non-missing entry in $V$ is randomly picked, and the partial gradients with regard to the involved $x_j$ and $y_i$ are calculated in order to update them. We further improve the optimization efficiency with asynchronous SGD, which updates parameters in parallel for multiple non-missing entries of $V$. Because the user-item view matrix is extremely sparse, there is little chance of parameter update conflict for our asynchronous SGD. In practice, we find that 
asynchronous SGD can significantly speed up the optimization without comprising the quality of the solution. We also find that it has a stable convergence 
behavior in practice. 

\begin{algorithm}[t!]
\caption{Negative Sampling}
\label{alg:sampling}
\textbf{Input:} initial view matrix $V\in \mathbb{R}^{M \times N}$, sampling probability $Pr_{ij}$, number of negatives to sample $n_j$ for each user   \\
\textbf{Output:} updated view matrix with negative sample set.
\begin{algorithmic}[1]
	\State $\mathcal{N} \gets \Phi$
        \For {$j=1,\ldots,N$}
    	\State $\mathcal{N} \gets$ sample $n_j$ negatives according to $Pr_{ij}$
	\State $ \forall i\in \mathcal{N}, V_{ij} = 0$
	\EndFor
\end{algorithmic}
\end{algorithm}

\subsection{Image feature learning with pseudo classes}

As we will see in the experiment section, the computed latent factors of our content items encode rich high-level visual and semantic information of the corresponding content items. Therefore, the learned latent factors can serve as a good source of supervision for learning meaningful features for our social media, i.e., image 
visual features in this work. Inspired by the recent work of deep convolutional neural network (DCNN) on large-scale ImageNet \cite{DBLP:conf/nips/KrizhevskySH12} being able to learn quite generic visual features, we use the latent factors to create pseudo classes for our content items and then apply DCNN to learn high level features in a supervised way. 
\paragraph{Learning with pseudo classes.} Even though could be noisy, the latent factors can largely reflect the semantic relationship between our content items. Although this source of information can be used in different forms to learn visual features, such as learning from sampled triplets, we resort to a more traditional supervised way by creating pseudo classes from the content items. Specifically, we first cluster the latent factor space into K clusters $\{a_{1}, ..., a_{K}\}$ with k-means. Then we create the pseudo classes $\{c_1, ..., c_K\}$ by partitioning the content items based on the cluster index of its latent factor, i.e., 
\begin{equation}
c_k = \{I_i:  y_i \in a_k\}. 
\end{equation}
Finally, we use the same deep neural network structure proposed in~\cite{DBLP:conf/nips/KrizhevskySH12}, which contains five convolutional layers and two fully connected layers, to learn a K-way DCNN classification model, from which we can extract high-level visual features for our social content.

Since the latent factor space is continuous, the above procedure might suffer from suspected quantization problem. Alternatively, we could learn a DCNN regression function directly from our content item to its latent factor \cite{van2013deep}, which defines a continues mapping between the visual images and latent factors. However, in practice, we find that training the network in this way results in inferior features. First, the latent space is a high-dimensional continuous space; learning a regression function directly from image pixels to this continuous space is very slow. Second, the latent factors obtained from the user behavior data is noisy and encode both visual semantic information and some amount of pure social and cultural information. Directly enforcing the mapping between image pixels and noisy latent factors could screw up the learning process, especially when we are using an $\ell_2$ norm cost function that is not robust to outliers. By creating discrete pseudo classes, which is essentially a lossy vector quantization coding transform, we could use the more robust softmax loss function that is robust to outliers. On the other hand, as the latent factors are noisy themselves, it may not matter much for the quantization loss. In our experiments, we find that learning from discrete pseudo classes can produce satisfactory visual features. 

\begin{figure*}
\centering
{\includegraphics[scale=0.25]{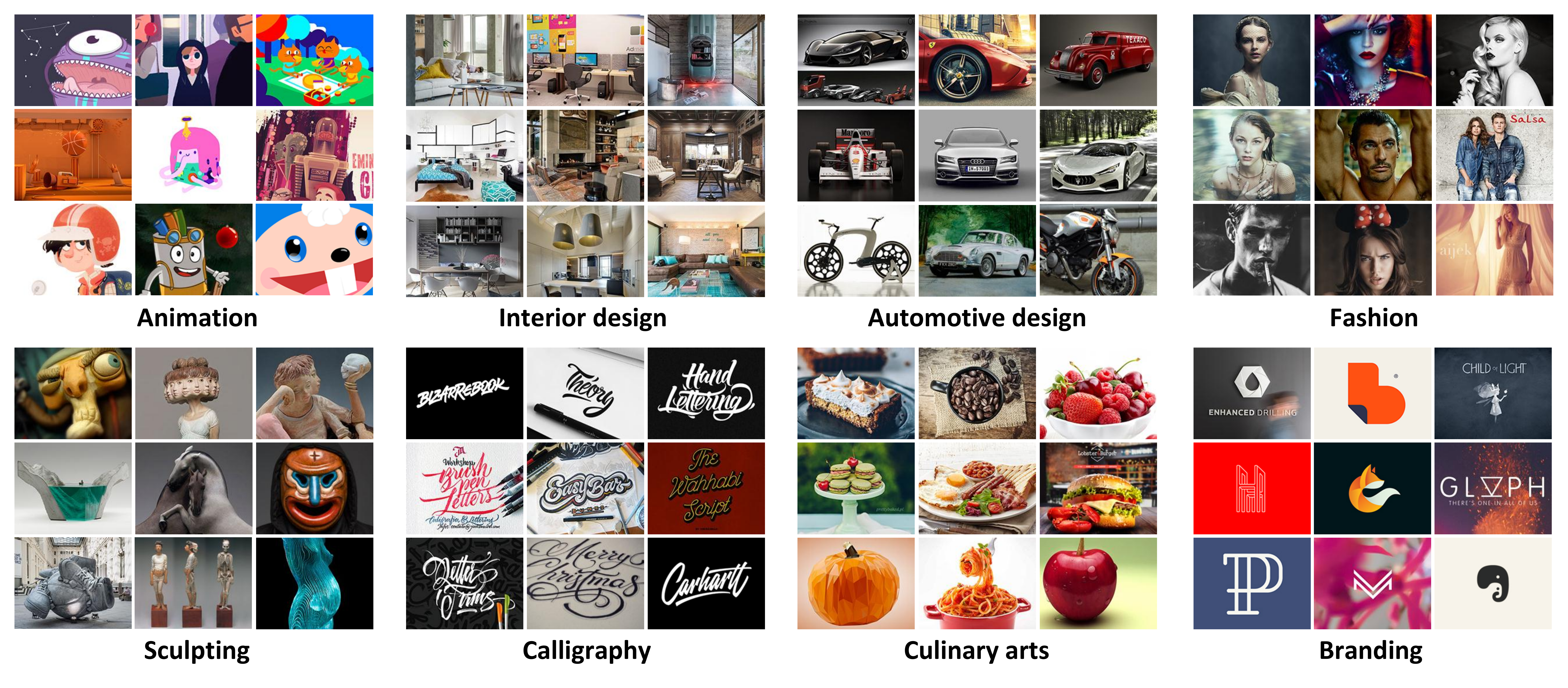}}
\caption{A subset of images from our Behance dataset. Images are
  organized according to the associated project fields, which are a very
  coarse categorization. Eight representative fields are
  shown. One may observe that the content on Behance is of high quality
  (professional photos, artwork, and designs) and very diverse (covering
  many different domains).}
\label{fig:db_repimage}
\end{figure*}

\section{Behance 2M Dataset}
To validate our new feature learning paradigm, we collect a large-scale
image and user behavior dataset from Behance.net. Behance is a popular
social media website for professional photographers, artists, and
designers to share their work. Content data on Behance are mostly in the
form of images and there are a small portion of videos as well. Content
is very diverse, ranging from photographs to cartoons, paintings,
typographs, graphic design, etc. Content data on Behance are organized
as projects, each of which has associated images and videos.  The
website organizes all its projects into 67 fields. Each project
may be associated with multiple fields. (Note that fields are very
coarse categorization and have large overlaps between each
other. Therefore, they are not suitable as labels for image
classification training.) In Figure~\ref{fig:db_repimage}, we show
several images from some representative fields.

The content data on Behance are all shared by the users. The project
owner, who uploads the project, picks one of the most representative
images as the cover image, which will be presented to other users.
While browsing over a large number of cover images, a user simply clicks
the cover image of interest and will be directed to the project page
with the entire content. Behance records the view data for each project
which is a list of users who have viewed the project. Choosing Behance
as our testbed is motivated by the observation that a user tends to view
projects of similar contents or styles, which is the key to our
approach.

We first download all the cover images for 1.9 million projects, and for
each project we obtain the list of users who have viewed it, which
results in 326 million view records from 1.9 million users. Note that we only download the cover images and use them in our experiments. But it is possible to obtain the additional project
content. The density of the view matrix (without negative sampling) is
about 0.0093\%.

We further process the raw data by removing the most popular and the
least popular projects, because projects with too many or too few views
cannot be modeled properly by latent factors. Similarly, we remove the
users that are too active and inactive. The minimum and maximum
thresholds we use for both project view counts and user view count are
(10, 20000). The thresholds are chosen to retain most of the data. In
particular, after thresholding, we have 1.9 million project and 310
million view records from 0.93 million users. The density of the view
matrix goes up to 0.0176\%. Although the most and least popular projects and active users have been removed, the distribution of views on projects, as well as the activity (the number of views a user gives) of users, are still uneven and long tail.



\section{Experiments}
In this section, we first study the characteristics of our learned latent factors and analyze the information captured. Then we investigate the neural network based visual feature, and evaluate its performance on Behance dataset for image similarity, style classification and image category classification on standard benchmarks.

\subsection{Experiment details}
\paragraph{Datasplit.} We split the processed data carefully for fair experiments. First, we split the set of all images as 95\% and 5\% as training and testing. For clarity, we will denote them as $tr$ and $te$ for the rest of the paper. This first split is for feature learning, thus $tr$ is used for DCNN training and $te$ is for evaluating learned feature representation. While training DCNN, we leave a small portion of $tr$ for validation purpose. Accordingly, the user behavior matrix after negative sampling $V$ is split based on $tr$ and $te$: we form matrix $V_{tr}$ by slicing $V$ with images belong to $tr$, and similarly we have $V_{te}$. Next, we split $V_{tr}$ for latent factor analysis. Note that, because the goal of latent factor analysis is to recover the missing entries in $V$, the split is on the non-missing entries in $V_{tr}$. We randomly sample 80\% of the non-missing entries for training, and the rest 20\% for validation. 

\paragraph{Latent factor learning.} To infer latent factors of projects in $tr$, we first sample negative entries for $V$ following Algorithm~\ref{alg:sampling}. For each user $j$, we set the number of negative samples, $n_j$, to be twice the number of her/his positives. Therefore, $V$ has about 900 million non-missing entries, so the matrix density becomes 0.0527\%. Note that as $n_j$ gets larger, the computational cost to solve Equation (\ref{eq:SVD}) increases linearly, thus for computational efficiency we avoid sampling too many negatives. We then apply regularized matrix factorization on the training split of $V_{tr}$, and use the validation split to estimate the optimal parameter value for $\lambda$. The validation criteria are root-mean-square error (RMSE) on predicting the validation split of $V_{tr}$ with inferred latent factors, and personalized ranking (PR), which measures the rank of positive and negative items for each user. Note that both RMSE and PR are used in an industrial machine learning competition~\cite{DBLP:journals/jmlr/DrorKKW12}. Then the entire $V_{tr}$ is used to compute final latent factors with the optimal $\lambda$. We also experimented with the dimensionality $d$ of $x$ and $y$. Generally, very high dimensional latent space (large $d$) may cause overfitting and increase computational cost, while very low ones may fail to capture the latent structure. Our validation process shows that setting $\lambda$ to 0.01 and $d$ to 100 is a reasonable balance which achieves good RMSE and PR, while keeping the computation efficient. With this setting, we obtain validation RMSE of 0.2955, and PR of 0.1891. Note that small values for both measurements means a good performance, and PR's expectation at random guess is 0.5. Please refer to~\cite{DBLP:journals/jmlr/DrorKKW12} for more details of RMSE and PR.

\paragraph{Image feature learning.} Following previous steps, we run k-means clustering on the learned latent factors. We vary the number of clusters within $[200, 500, 1000, 2000, 3000, 5000]$, and train a DCNN for each of the the corresponding cluster assignments. Similar to the data preparation pipeline in Krizhevsky et al.~\cite{DBLP:conf/nips/KrizhevskySH12}, we resize training images so that the short side has 256 pixels, then take the center crop of the resized images. At training time, we use random $224 \times 224$ crops to augment the dataset for improved robustness. Dropout is adopted to avoid overfitting. Training approximately converged in 60 epochs, and we extract features from fully connected layers in DCNNs.

\begin{figure*}[t!]
\centering
\hspace{-5.5mm}\includegraphics[scale=0.38]{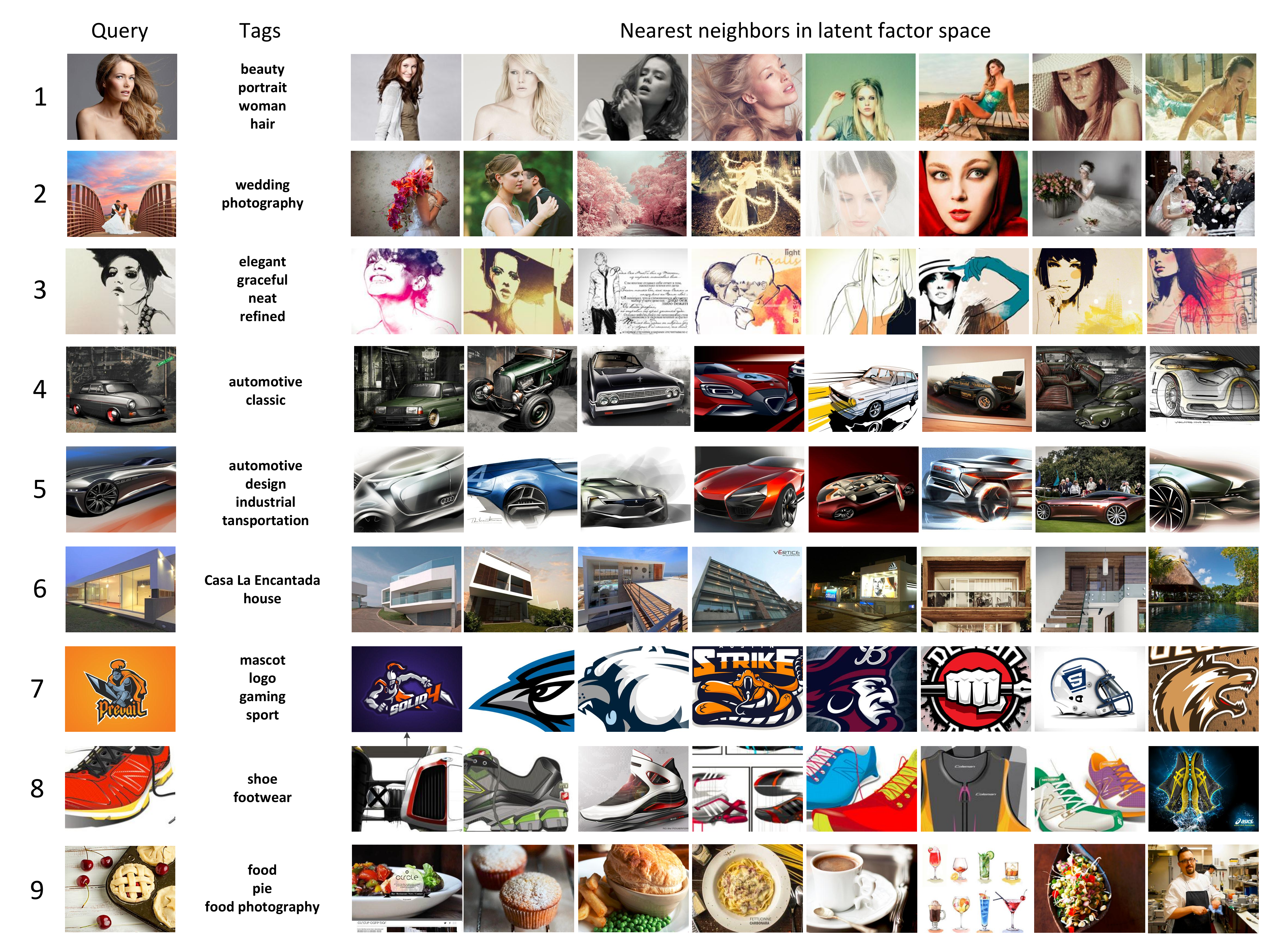}
\caption{Query images and their retrieved nearest neighbors (NNs) in latent factor space. Queries and their NNs share similar visual semantics and contexts. Both coarse and fine level semantics are well captured (compare Row 1-3 and Row 4-5). For easy illustration, representative tags of query images are listed as well.}
\label{fig:latent}
\end{figure*}

\subsection{Analysis of latent factors}

As discussed earlier, latent factors from view data should reveal some properties of the content data. Since the latent factors also serve as an implicit supervision in our feature learning, it is natural to learn what information they have captured about the individual images and whether our assumption of the correlation structure is valid. To this end, we present a simple experiment to empirically study those latent factors. For each project, represented by its cover image, we retrieve its nearest neighbors in the training set $tr$, using cosine similarity between latent factors. We find strong visual and semantic proximity between query projects and their nearest neighbors (NNs), and the observation is consistent across the entire set. In Figure~\ref{fig:latent}, We show several randomly selected queries of various semantics and their top NNs. For clear illustration, we also list representative tags of the queries. Note that tags are obtained from Behance, and they are provided by project owners at upload time, therefore, tags are typically random and noisy. (We only show the informative ones here.)

From Figure~\ref{fig:latent}, we can first observe a clear categorical correlation at coarse level between a query and its NNs across all examples. For instance, row 1-3 are all portraits of woman, row 4-5 are about automotive design, and the following rows include various subjects and contexts, such as house, footwear, food photo etc. Furthermore, we notice that the latent factors also reveal richer visual context at a finer level than category. For instance, despite all being female portraits, query 1-3 have drastically different image styles and contexts, and these differences are well respected in the retrieved NNs.
Similar phenomenon is shown in row 4 and 5, where the car in query 4 is a classic car, thus there are more classic cars in its NNs while cars in row 5 are more of modern design. On the other hand, we also observe a small portion of failed queries, whose NNs are irrelevant. These failure cases are mainly caused by two reasons: (1) the sparsity and noise in user behavior data, (2) social factors that are not visually related, e.g. an user always like to view his/her friends' projects.



\subsection{Image similarity on Behance}
Given the latent factors analyzed previously, a natural question is how well our learned image feature 
captures the concept embedded in the latent space. To validate this, we conduct a retrieval experiment on the learned image features. We use images in $te$ to query against $tr$. Similarity between images are calculated using cosine similarity. To validate the effectiveness of our learned feature, we compare it with the state-of-the-art image classification feature learned on ImageNet ILSVRC2012 dataset, with the same network structure and training procedure as~\cite{DBLP:conf/nips/KrizhevskySH12}, and we refer to it as ImageNet feature.

\begin{figure}[t!]
\centering
\includegraphics[scale=0.25]{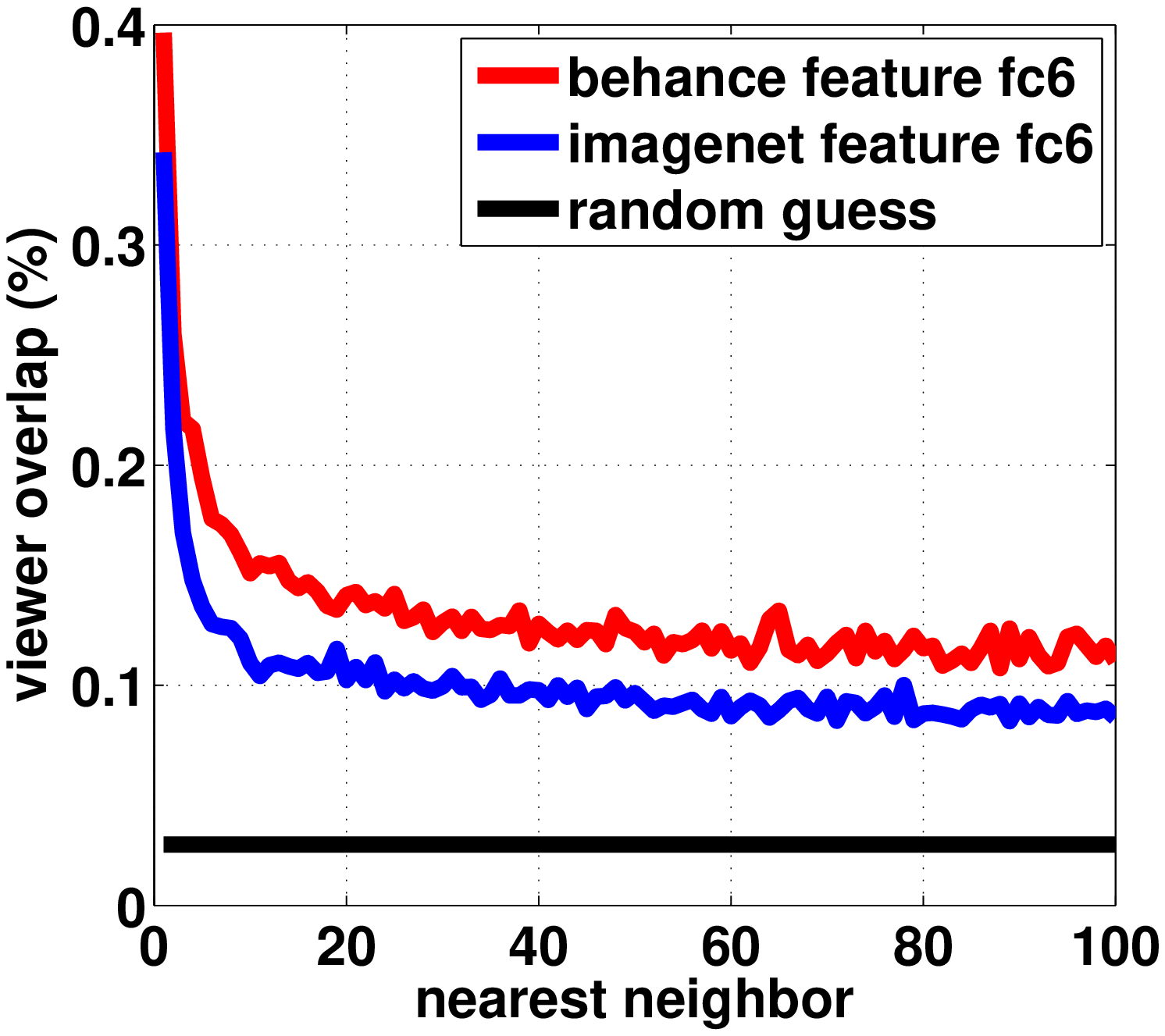}
\includegraphics[scale=0.25]{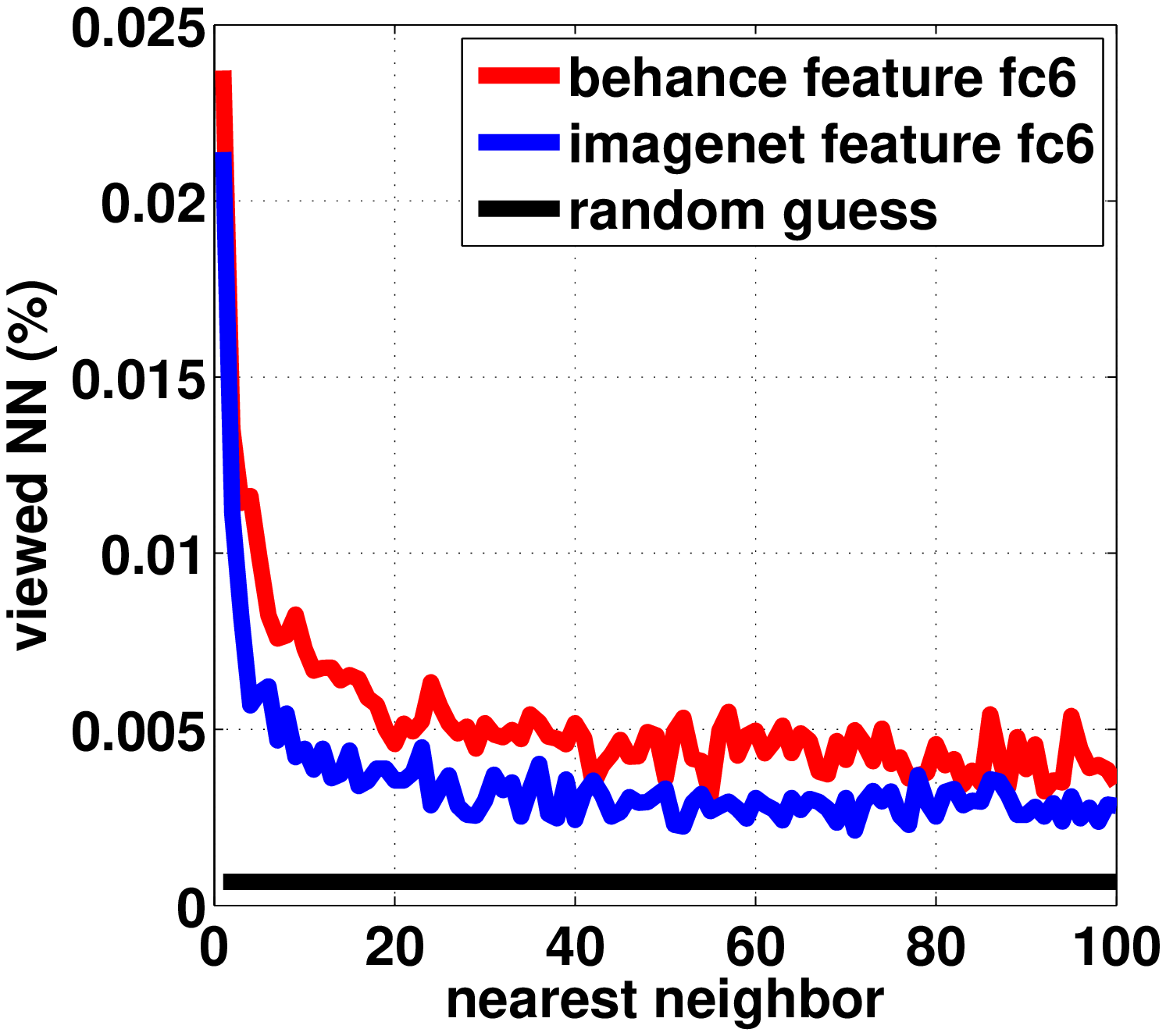}
\caption{Quantitative comparison of our feature and ImageNet feature for retrieval task on Behance images. Our feature performs better in both measurements. The black line is the value of the measurements obtained by random sampling image pairs (random retrieval and random NNs).}
\label{fig:exp1}
\end{figure}

We first visually inspect the visual relationship between a large number
of queries and their NNs. In Figure~\ref{fig-feature-compare}, we show a
set of randomly selected queries and the NNs for our feature and the
ImageNet feature \cite{DBLP:conf/nips/KrizhevskySH12}, and clearly our feature shows significantly better
visual consistency.

Furthermore, we quantitatively evaluate the NNs with two measurements, which are designed for Behance data. The measurements go beyond visual similarity, and most importantly, reflect the relationships between images on Behance. The measurements are: given a query image, 

(1) the number of common viewers between the query and the retrieved NNs. We measure, for each pair of query and NN, the ratio between their common viewers (set size) and the union (set size), therefore, as shown in Figure~\ref{fig:exp1}, the ratio drops as the rank of neighbors falls further behind. We calculate this measure for the top 100 NNs for each query in $te$, and report its mean across $te$ at every rank position;

(2) the number of retrieved NNs that have been viewed by the owner of the query image. Similarly, this quantity is measured between a query and all of the top 100 NNs. We report the average of this number across all queries in $te$.


Plots on the two measurements are shown in Figure~\ref{fig:exp1}, where our feature consistently outperforms ImageNet feature. Because of marginal differences of DCNN features from Fully Connected Layer (FC) 6 and FC7 layers, we report the results of FC6 features. (For better comparison, we also randomly sample a large number of image pairs and calculate the random expectation of the two measurements.)

\subsection{Image classification on Benchmarks}
We apply our feature as image descriptor to image classification on standard benchmarks for object class classification and visual style classification. We choose Caltech256~\cite{griffin2007caltech} for object class classification. We focus more on image style classification, due to the nature of our dataset, which is crawled from an artistic asset sharing site. We use three major image style benchmarks, including Flickr style, which has 80k images on 20 visual styles, Wikipaintings, which contains 85k images belonging to 25 styles, and AVA style, which consists of 14k images with 14 photographic styles. Flickr style and Wikipaintings are both from a recent work~\cite{karayev2013recognizing}, in which Karayev et al. evaluates various established image features on the three benchmarks.

\begin{table}[t!]
\small
\caption{Classification accuracy on benchmarks (\%)} 
\centering 
\begin{tabular}{c c c c}
\hline\hline 
&Our feature&ImageNet feature&Meta-Class\\ 
\hline
Flickr style&\textbf{37.2}&37.1&32.8\\ 
Wikipaintings&\textbf{41.4}&40.7&38.6\\
AVA style&\textbf{56.0}&51.3 &53.9\\
Caltech256&57.6&\textbf{68.9} &48.9\\
\hline 
\end{tabular}
\label{table:nonlin}
\end{table}

We select ImageNet feature~\cite{DBLP:conf/nips/KrizhevskySH12} and Meta-Class~\cite{BergamoTorresani:CVPR2012} as competing features for their competitive performance. For deep learning features, we have experimented with features from both FC6 and FC7 layers, and found that for object classification, the performance difference is marginal, however, for image style classification, due to dataset bias, features in FC7 layer obtains lower accuracy. Therefore, we report results on FC6 features. We also experimented with our features trained on different number of pseudo classes, and found that 2000 consistently provides better performance than others.  Each of the style benchmarks is split into 80\% and 20\% for training and testing, and for Caltech256, we use 50 images per category for training and 20 images for testing. We use linear SVM as the classification model.

As shown in Table \ref{table:nonlin}, in image style classification benchmarks, our feature achieves similar or even better accuracy than ImageNet feature, which is the state-of-the-art single feature for style recognition reported in~\cite{karayev2013recognizing}. As for object classification, our feature obtained competitive results on Caltech256. It is worth to point out that ImageNet feature is learned on ILSVRC2012 dataset with 1000 categorical labels, and Met-Class is trained on a subset of 8000 synsets of the entire ImageNet database, whereas our feature is learned on images with noisy user view data.

\section{Discussion}



We propose a novel data-driven feature learning paradigm that exploits user behavior data collected from social media websites. Our feature learning paradigm is different from existing methods in that it does not rely on category labels. We show that the learned feature outperforms the state-of-the-art features on our Behance 2M Dataset in terms of learning better image similarities while performing competitively on various standard recognition benchmarks. We believe using social data for feature learning and image recognition in  general is a promising new research direction. 

In terms of future work, we plan to pursue two directions. First, we want to make our learning paradigm more robust to data noise and non-visual factors. Second, we would like to go beyond the view records and incorporate additional social data, such as user relationships on social media. For more information, please visit the project webpage\footnote{\url{http://www.cs.dartmouth.edu/~chenfang/proj_page/Behance_CVPR15/}}.

{\small
\bibliographystyle{ieee}
\bibliography{main_arvix}
}

\end{document}